\documentclass[a4paper]{article}
\pdfoutput=1

\usepackage{spconf}
\usepackage{graphicx}
\usepackage{amsmath,bm}
\usepackage{textcomp}
\usepackage{xcolor}
\usepackage{color}
\usepackage{url}
\usepackage{epstopdf}
\usepackage{footnote}

\def\vec#1{\ensuremath{\boldsymbol{{#1}}}}

\ninept

\newcommand{\seqtoseq}{\protect{encoder-decoder}}
\newcommand{\Seqtoseq}{Encoder-decoder}
\newcommand{\pron}[1]{[ #1 ]}

\title{Jointly learning to align and convert graphemes to phonemes \\ with neural attention models}
\name{Shubham Toshniwal, Karen Livescu}
\address{Toyota Technological Institute at Chicago\\
 {\small \tt \{shtoshni, klivescu\}@ttic.edu}
 }
\begin{document}
\maketitle
\begin{abstract}
We propose an attention-enabled \seqtoseq{} model for the problem of grapheme-to-phoneme conversion.
Most previous work has tackled the problem via joint sequence models that require explicit alignments for training. In contrast, the attention-enabled \seqtoseq{} model allows for jointly learning to align and convert characters to phonemes. 
We explore different types of attention models, including global and local attention, and our best models achieve state-of-the-art results on three standard data sets (CMUDict, Pronlex, and NetTalk). 
\end{abstract}

\noindent{\bf Index Terms}: grapheme-to-phoneme, LSTM, \seqtoseq, attention

\section{Introduction}
Grapheme-to-phoneme conversion (``G2P'') is the task of converting from a spelling (a grapheme sequence) to its pronunciation (a phoneme or phone sequence\footnote{We will generally use the term ``phoneme'' to encompass any subword unit of pronunciation.}). 
For example, given the word \textit{paste}, the task is to output its pronunciation \pron{P EY S T}.
This conversion is a frequent component of text-to-speech (TTS) and automatic speech recognition (ASR) systems.
While static dictionaries exist, they are finite; G2P models are essential for handling new words.

One of the challenges in G2P conversion is that the pronunciation of any grapheme depends on a variety of factors including its context and the etymology of the word.
For example, \textit{defiance} is pronounced \pron{D IH F AY AH N S}, while the similar grapheme sequence \textit{fiance} is pronounced in a very different way \pron{F IY AA N S EY} because of its origin in French.  
Another complication is that the output phone sequence can be either shorter than or longer than the input grapheme sequence.  For example, \textit{knife} $\longrightarrow$ \pron{N AY F} has two ``silent'' graphemes at the beginning and end, while \textit{exit} $\longrightarrow$ \pron{EH G Z IH T} has more phones than graphemes.  The alignment between grapheme and phoneme sequences is therefore non-trivial.

A typical approach to G2P involves using joint sequence models, where the alignment is provided via some external aligner \cite{Chen, Bisani, m2malignment}. However, 
since the alignment is a latent varible---a means to an end rather than the end itself---it is interesting to consider whether we can do away with such explicit alignments. 

Some recent work on the G2P problem has used neural network-based approaches.  Specifically, long short-term memory (LSTM) networks~\cite{hochreiter1997lstm} have recently been explored \cite{Rao, Yao}.  LSTMs (and, more generally, recurrent neural networks) can model varying contexts (``memory'') and have been successful for a number of sequence prediction tasks.  When used in an encoder-decoder approach, as in \cite{Yao}, they in principle require no alignment between the input (grapheme sequence) and output (phoneme sequence) and are therefore quite natural for this task.  However, to date the best-performing G2P approaches still use alignments.

In this paper we explore an extension of encoder-decoder networks based on an attention mechanism, which has proven useful in other sequence prediction tasks~\cite{bahdanau,LAS,SAT}.
The attention mechanism endows encoder-decoder networks with the ability to consider ``soft'' alignments, and to learn these alignments jointly with the sequence prediction task.  We show that we can improve over the state-of-the-art G2P results on three standard data sets with such attention-enabled \seqtoseq{} models, removing entirely the dependency on alignments.\footnote{Code available at \url{https://github.com/shtoshni92/g2p}}

\section{Related Work}
%The G2P problem can be viewed as a specific instance of machine translation problem with the source language being grapheme based and the target language being phoneme based.
Most previous work on learning G2P breaks the problem into two steps: (i) align the grapheme and phoneme sequences (ii) train a conditional or joint maximum entropy model on the aligned data \cite{Chen, Bisani, m2malignment, Galescu}. For example, an alignment between the grapheme sequence \textit{phoneme} and the phoneme sequence \pron{F OW N IY M} might be as follows:
\begin{table}[hbt]
\centering
\begin{tabular}{cccccc}
ph & o & n & e & m & e\\
\textbar & \textbar & \textbar & \textbar & \textbar & \textbar \\
f & ow & n & iy & m & - \\
\end{tabular}
%\caption{Alignment between the grapheme and phoneme sequence for word ``phoneme"}
\label{tab:align}
\end{table}

Given a grapheme sequence (which has been aligned/chunked) $G=g_1, \cdots, g_m$, conditional models estimate the probability $\Pr(P|G)$ of a phoneme sequence $P = p_1, \cdots, p_m$ by:
$$\Pr(P|A, G) \approx \prod_{i=1}^{m}\Pr(p_i | p_{i-k}^{i-1}, g_{i-k}^{i+k})$$
for some appropriate alignment $A$ and a context window of size $k$. The factored terms in the distribution can be modeled using a maximum entropy classifier \cite{Chen}, or the full product can be modeled as a conditional random field (CRF) \cite{Wang}.

Another approach is to do joint sequence modeling by constructing a vocabulary of aligned grapheme and phoneme pairs, or \emph{graphones}. The graphone sequence can be modeled via $n$-gram models \cite{Bisani, Galescu} or maximum entropy models \cite{Chen}.

Recently there has been some work using recurrent neural networks (RNNs) for this task. Rao {\it et al.}~\cite{Rao} use bidirectional LSTMs with a connectionist temporal classification (CTC) layer~\cite{CTC} which doesn't require the data to be aligned.  Yao and Zweig~\cite{Yao} explore both an encoder-decoder architechture \cite{seq2seq} and an input-output LSTM network with explicit alignments. Among these previous approaches, the best performance by a single model is obtained by Yao and Zweig's alignment-based approach, although Rao {\it et al.} obtain even better performance on one data set by combining their LSTM model with an (alignment-based) $n$-gram model.

In this paper, we explore the use of attention in the encoder-decoder framework as a way of removing the dependency on alignments.  The use of a neural attention model was first explored by Bahdanau {\it et al.} for machine translation~\cite{bahdanau} (though a precursor of this model was the windowing approach of Graves~\cite{graves2013generating}), which has since been applied to a variety of tasks including speech recognition~\cite{LAS} and image caption generation~\cite{SAT}.  The G2P problem is in fact largely analogous to the translation problem, with a many-to-many mapping between subsequences of input labels and subsequences of output labels and with potentially long-range dependencies (as in the effect of the final ``e'' in \textit{paste} on the pronunciation of the ``a'').  In experiments presented below, we find that this type of attention model indeed removes our dependency on an external aligner and achieves improved performance on standard data sets.

\section{Model}
We next describe the main components of our models both without and with attention.

\begin{figure}[hbt]
\centering
\includegraphics[width=0.5\textwidth]{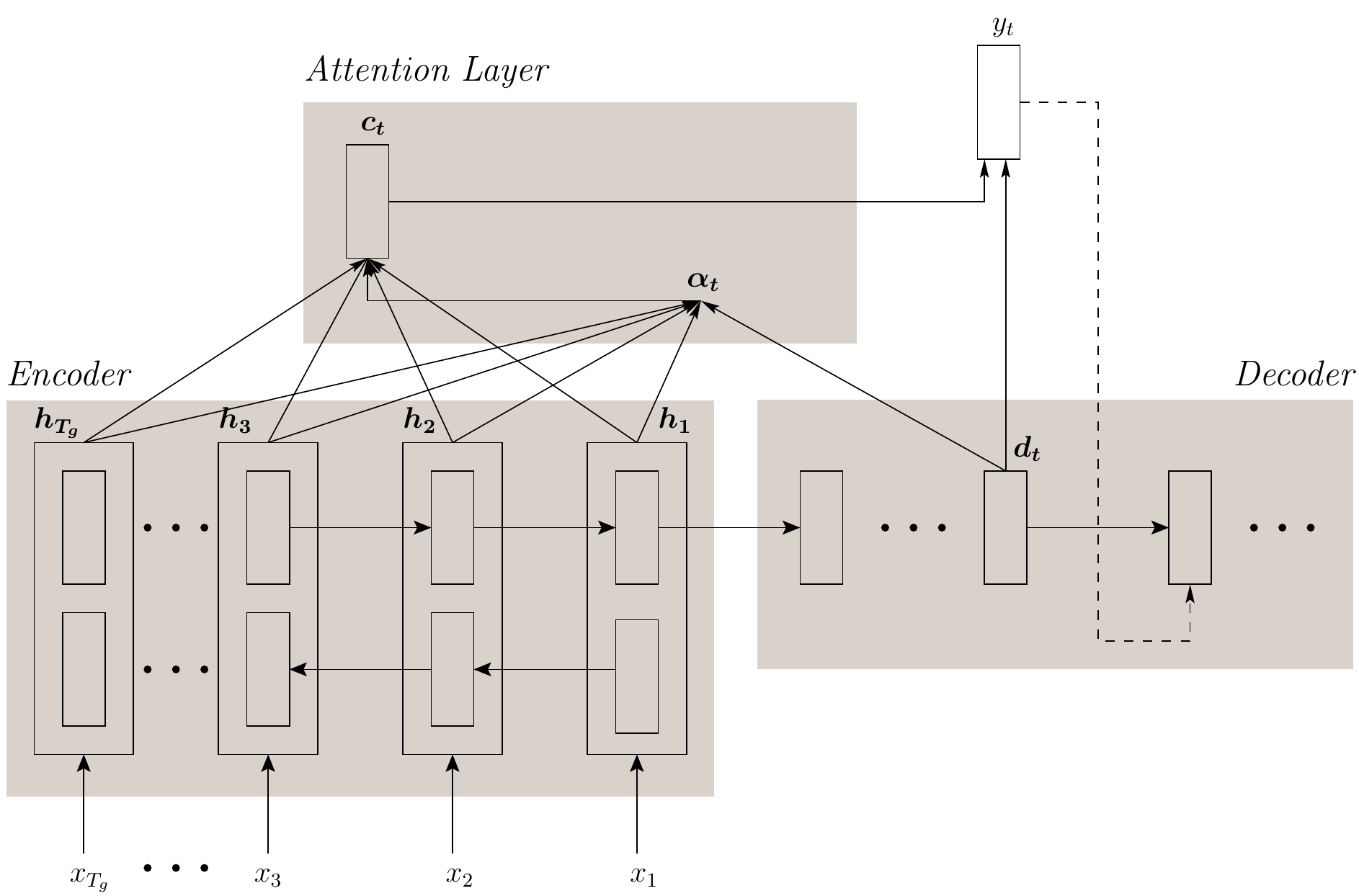}
\caption{A global attention encoder-decoder model reading the input sequence $x_1, \cdots, x_{T_g}$ and outputting the sequence $y_1, \cdots, y_t, \cdots$}
\label{fig:encdec}
\end{figure}
\vspace*{-0.3in}
\subsection{\Seqtoseq{} models}
We briefly describe the \textit{\seqtoseq} (``sequence-to-sequence'') approach, as proposed by \cite{seq2seq}.  An \seqtoseq{} model includes an encoder, which reads in the input (grapheme) sequence, and a decoder, which generates the output (phoneme) sequence. A typical \seqtoseq{} model is shown in Figure \ref{fig:encdec}. In our model, the encoder is a bidirectional long short-term memory (BiLSTM) network; we use a bidirectional network in order to capture the context on both sides of each grapheme.  The encoder takes as input the grapheme sequence, represented as a sequence of vectors $\vec{x} = (\vec{x}_1, \cdots, \vec{x}_{T_g})$, obtained by multiplying the one-hot vectors representing the input characters with a character embedding matrix which is learned jointly with the rest of the model.  The encoder computes a sequence of hidden state vectors, $\mathbf{h} = (\vec{h}_1, \cdots, \vec{h}_{T_g})$, given by:
\begin{align*}
\overrightarrow{\vec{h}_i} &= f(\vec{x}_i, \overrightarrow{\vec{h}_{i - 1}})\\
\overleftarrow{\vec{h}_i} &= f'(\vec{x}_i, \overleftarrow{\vec{h}_{i + 1}})\\
\vec{h}_i &= (\overrightarrow{\vec{h}_i}; \overleftarrow{\vec{h}_i})
\end{align*}
\noindent We use separate stacked (deep) LSTMs to model $f$ and $f'$.\footnote{For brevity we exclude the LSTM equations.  The details can be found in Zaremba {\it et al.} ~\cite{lstm_eqn}.}
A ``context vector'' $\vec{c}$ is computed from the encoder's state sequence:
$$\vec{c} = q(\{\vec{h}_1, \cdots, \vec{h}_{T_g}\})$$
In our case, we use a linear combination of $\overrightarrow{\vec{h}_{T_g}}$ and 
$\overleftarrow{\vec{h}_{1}}$, with parameters learned during training. Since our models are stacked, we carry out this linear combination at every layer.

This context vector is passed as an input to the decoder.  The decoder, $g(\cdot)$, is modeled as another stacked (unidirectional) LSTM, which predicts each phoneme $y_{t}$ given the context vector $\vec{c}$ and all of the previously predicted phonemes $\{y_1, \cdots, y_{t - 1}\}$ in the following way:
\begin{align*}
 \vec{d}_t &= g(\vec{\tilde{y}}_{t-1}, \vec{d}_{t-1}, \vec{c})\\
 p(y_{t} | y_{< t}, \vec{x}) &= \text{softmax}(\vec{W_s} \vec{d}_t + \vec{b_s})
\end{align*}
where $\vec{d}_{t-1}$ is the hidden state of the decoder LSTM and $\vec{\tilde{y}}_{t-1}$ is the vector obtained by projecting the one hot vector corresponding to $y_{t-1}$ using a phoneme embedding matrix $\vec{E}$. The embedding matrix $\vec{E}$ is jointly learned with other parameters of the model. In basic \seqtoseq{} models, the context vector $\vec{c}$ is just used as an initial state for the decoder LSTM, $\vec{d}_0 = \vec{c}$, and is not used after that. 
\subsection{Global Attention}
One of the important extensions of \seqtoseq{} models is the use of attention mechanism to adapt the context vector $\vec{c}$ for every output label prediction~\cite{bahdanau}.  Rather than just using the context vector as an initial state for the decoder LSTM, we use a different context vector $\vec{c}_t$ at every decoder time step, where $\vec{c}_t$ is a linear combination of all of the encoder hidden states.
%This access to all the encoder hidden states means that there are now multiple paths of back-propagating error and hence,
The choice of initial state for the decoder LSTM is now less important; we simply use the last hidden state of the encoder's backward LSTM.
The ability to attend to different encoder states when decoding each output label means that the attention mechanism can be seen as a soft alignment between the input (grapheme) sequence and output (phoneme) sequence. 
We use the attention mechanism proposed by \cite{Vinyals}, where the context vector $\vec{c}_t$ at time $t$ is given by: 
\begin{align*}
u_{it} &= \vec{v}^T \text{tanh}(\vec{W_1}\vec{h}_i + \vec{W_2}\vec{d}_t + \vec{b_a}) \\
\vec{\alpha}_{t} &= \text{softmax}(\vec{u}_t) \\
\vec{c}_t &= \sum_{i=1}^{T_g}\alpha_{it}\vec{h}_i 
\end{align*}
\noindent where the vectors $\vec{v}, \vec{b_a}$ and the matrices $\vec{W_1}, \vec{W_2}$ are parameters learned jointly with the rest of the encoder-decoder model.
The score  $\alpha_{it}$ is a weight that represents the importance of the hidden encoder state $\vec{h}_i$ in generating the phoneme $y_t$. It should be noted that the vector $\vec{h}_i$ is really a stack of vectors and for  attention calculations we only use its top layer.

The decoder then uses $\vec{c}_t$ in the following way: 
\begin{align*}
p(y_t | y_{< t}, \vec{x}) &= \text{softmax}(\vec{W_s} [\vec{c}_t; \vec{d}_t] + \vec{b_s})
\end{align*}

\subsection{Local Attention}
The global attention mechanism considers all of the encoder states in predicting each decoder output.  It can be argued, however, that the G2P problem has an inherently local nature; that is, each output typically depends on only a small window of input labels.  Therefore, limiting attention to an \emph{appropriate} window of encoder states should be sufficient to predict each phoneme.  \emph{Local attention} models operate by first finding an aligned position $p_t$ at each timsestep $t$, which is an index into the input sequence, and then considers a context window [$p_t - D$, $p_t + D$] of indices around it.  In our experiments, we find $D=3$ to work well.  The context vector $c_t$ is then calculated in a similar fashion as in the global attention model described above but while considering only the encoder states falling within the context window.  We consider two variants of local attention, based on \cite{Luong}, described below.
\subsubsection{Monotonic Alignment (local-$m$)}
The local-$m$ model makes the assumption of a simple monotonic alignment between the source and target sequence: $p_t = t$.  The attention weights are then computed as in the global attention model, but the context vector is computed by summing only over [$p_t - D$, $p_t + D$].  This type of attention model is a reasonable one for tasks where the input and output sequences can be expected to proceed almost in lock step, which is arguably the case in the G2P task.
\subsubsection{Predictive Alignment (local-$p$)}
The local-$m$ model may be too restrictive.  An alternative is to include the prediction of the attention center $p_t$ as part of the model.  The local-$p$ model predicts $p_t$ in the following way:
$$p_t = T_g  \cdot \sigma(\vec{v_{p}}^T \text{tanh}(\vec{W_p} \vec{d}_t)) $$
\noindent where $T_g$ is the length of the input sequence, $\sigma(\cdot)$ is a logistic function, and $\vec{v_{p}}, \vec{W_p}$ are learned parameters.  The $\sigma(\cdot)$ ensures that the predicted index does not exceed the source length.  In addition, the model also favors the input positions near $p_t$ by reweighting the attention weights with a Gaussian prior centered around $p_t$. The new attention weights $\widetilde{\alpha_{it}}$ are then given by:
$$\widetilde{\alpha_{it}} = \alpha_{it} \cdot \text{exp}\left(- \frac{(i - p_t)^2}{2{\sigma}^2}\right)$$
for encoder positions $i$ falling within the context window, where $\alpha_{it}$ is as computed for the global attention model and the Gaussian standard deviation $\sigma = \frac{D}{2}$.  Finally, the new attention weights $\widetilde{\alpha_{it}}$ are used in place of $\alpha_{it}$ in computing the context vector.
\subsection{Input Feeding}
One of the issues with attention models is that at each decoder step, the attention model decides what to attend to independently of the decisions at other time steps.  Therefore, there is no explicit global constraint to ensure that all of source sequence is ``covered", i.e.~that all of the inputs are being attended to at one point or another.  One indirect way of addressing this potential problem is by concatenating the previous attention-weighted context vector with the input for the next decoder time step, as done in \cite{Luong}.  This ``feeding" of previous attention makes the model more aware of the attention decisions being made in previous time steps.  We treat the choice of using input feeding as a hyperparameter of our models.

\section{Experiments} 
\subsection{Data}
In order to compare directly with earlier results, we use three standard data sets typically used for evaluating G2P models, namely  CMUDict, Pronlex and NetTalk. For all three data sets, we use the same experimental setup as in~\cite{Chen}, which we briefly describe below.\footnote{We are grateful to Stan Chen for providing the data.}

The CMUDict data set is split into a 106,837-word training set and a 12,000-word test set. As in previous work, we sample 2,670 words from the training set to create a development set for the purpose of hyperparameter tuning.

The Pronlex data set has 83,182 words in the training set, 2,400 in the development set, and 4,800 in the test set.  The development and test sets are split into 3 categories.  We report only the overall results and not category-wise ones for Pronlex.

The NetTalk data set is quite small compared to the other two data sets, with only 14,851 words for training and 4,951 words for testing. For the purpose of hyperparameter tuning, we create a development set of 1,000 words sampled from the training set; however, after tuning, the final model is trained on the original training set. 

\subsection{Evaluation}
We evaluate performance using the standard measures of word error rate (WER) and phoneme error rate (PER), reported as percentages.
PER is equal to the Levenshtein distance of the predicted phoneme sequence from the ground truth divided by the total number of phonemes in the ground truth. 
WER is equal to the total number of word in which there is at least one phone error, divided by the total number of words.
As in prior work, for words with multiple ground-truth pronunciations, we choose the ground truth that results in the lowest PER.

\subsection{Training}
Our stacked LSTMs have 3 layers, each with 512 units. We use 512-dimensional embedding vectors for representing both the characters and the phonemes. We use minibatch stochastic gradient descent (SGD) together with Adam~\cite{adam} using a minibatch size of 256. We use an initial learning rate of 0.001 and reduce this learning rate by a multiplicative factor of 0.8 whenever the WER for development data does not decrease at the end of an epoch. We train the model for 100 epochs but update/save it at the end of an epoch only when it decreases the WER on development data. To prevent overfitting we: (a) introduce a dropout~\cite{srivastava2014dropout} layer between every pair of consecutive layers of the stacked LSTMs, and (b) use scheduled sampling \cite{sampling}, with a linear decay, on the decoder side. 

We tune the dropout probability over the range $\{0, 0.1, 0.2, 0.3, 0.4\}$.  We also tune over the choice of using input feeding or not. The combination of hyperparameters that gives the best result on the development set is chosen. It should be noted that the performance is quite stable with respect to dropout and input feeding:  We observed only a minor drop in performance even if we choose slightly ``sub-optimal" parameters. All of the models are implemented using TensorFlow \cite{tensorflow}.

\begin{savenotes}
\begin{table*}[hbt]
\centering
\begin{tabular}{| l | l | l | l | }
\hline
\textbf{Data} & \textbf{Method} & \textbf{PER (\%)} & \textbf{WER (\%)} \\ \hline
CMUDict & BiDir LSTM + Alignment \cite{Yao} & 5.45  &  23.55 \\ 
 & DBLSTM-CTC \cite{Rao}  & - & 25.8\\
 & DBLSTM-CTC + 5-gram model \cite{Rao} & - & $ 21.3$ \\

& \Seqtoseq{} + global attn & $5.04\pm0.03$ & $21.69\pm0.21$ \\
& \Seqtoseq{} + local-$m$ attn & $5.11\pm0.03$ & $21.85\pm0.21$   \\
& \Seqtoseq{} + local-$p$ attn & $5.39\pm0.04$ & $22.83\pm0.22$   \\
& Ensemble of 5 [\Seqtoseq{} + global attn] models & $\mathbf{4.69}$ & $\mathbf{20.24}$ \\\hline

Pronlex & BiDir LSTM + Alignment \cite{Yao} & 6.51  &  26.69 \\
        &\Seqtoseq{} + global attn & $6.24\pm0.1$ & $25.39\pm0.61$ \\
		& \Seqtoseq{} + local-$m$ attn & $\mathbf{5.99\pm0.11}$ & $\mathbf{24.23\pm0.42}$ \\
		& \Seqtoseq{} + local-$p$ attn & $6.49\pm0.06$ & $25.64\pm0.42$   \\\hline

NetTalk & BiDir LSTM + Alignment \cite{Yao} & $7.38$  &  30.77 \\ 

 		& \Seqtoseq{} + global attn & $\mathbf{7.14\pm0.72}$ & $\mathbf{29.20\pm2.18}$ \footnote{The high variance is due to 1 of the 5 runs being unusually bad, resulting in a WER of 33.47\%}\\%\protect\footnotemark\\
		& \Seqtoseq{} + local-$m$ attn & $7.13\pm0.11$ & $29.67\pm0.49$ \\
		& \Seqtoseq{} + local-$p$ attn & $8.41\pm0.19$ & $32.32\pm0.41$ \\\hline
\end{tabular}

\caption{Comparison of our models' performance with the best previous results for CMUDict, Pronlex and NetTalk. $\pm$ indicates the standard deviation across 5 training runs of the model.}
\label{tab:cmu}
\end{table*}
\end{savenotes}

\subsection{Inference}
We use a greedy decoder (beam size = 1) to decode the phoneme sequence during inference.  That is, at each decoding time step we consider the output phone to be the argmax of the softmax output of the decoder at that time frame.  We got no reliable gains by using beam search with any beam size greater than 1.

\subsection{Results}

Table~\ref{tab:cmu} presents the main results with our tuned models on the three test sets, compared to the best previously reported results.  For all three data sets, the best prior results to our knowledge with a single model are those of Yao and Zweig~\cite{Yao} with an alignment-based deep bidirectional LSTM.  For CMUDict, a better WER was obtained by~\cite{Rao} by ensembling their CTC bidirectional LSTM with an alignment-based 5-gram model, but no corresponding PER was reported.

Our best attention models clearly outperform all of the previous best models in terms of PER.  In terms of WER, the attention model outperforms all prior single (non-ensembled) models.  For CMUDict, we also include the result of ensembling five of our global attention models using different random initializations, by voting on their outputs (with random tie-breaking), which outperforms the ensemble of Rao {\it et al.}.

Among the three attention models, global and local-$m$ attention model perform well across all three data sets, while the local-$p$ model performs well on CMUDict and Pronlex but not on NetTalk. The success of the global model may be explained by the fact that the source sequence length in this task (i.e., the word length) is rather short, always less than 25 characters in these three data sets.  Therefore, it is feasible for the global attention model to consider all of the encoder states and weight them in an appropriate manner. 

The local-$m$ attention model, even with its simplistic assumption about alignment, outperforms the local-$p$ model on every data set and is a clear best performer on Pronlex.  Although the assumption of monotonic alignment turns out to be too simplistic for other tasks, such as machine translation \cite{Luong}, it is a reasonable choice for G2P.

Surprisingly, the local-$p$ attention model remains a distant third among the three attention models across all three data sets. Moreover, it suffers a higher PER even when it obtains comparable WERs, as is the case for Pronlex.  This means that words with errors tend to have a large number of errors.  This seems to suggest that if an alignment error is made near the beginning of the word, then it is hard for the local-$p$ model to recover.  This points towards a need for a better alignment prediction strategy. The particular poor performance of local-$p$ on NetTalk also suggests that it may need a larger training set for learning the alignment prediction parameters.

For all of our results, we make the choice of dropout probability and input feeding based on tuning on the development set. The typical choice of dropout probability tends to be around 0.2-0.3 while the decision of using input feeding or not tends to vary with the choice of attention model and the data set used.  However, performance does not vary greatly with these two parameters.

\subsubsection{Ablation analysis for CMUDict}
\begin{table}
\begin{tabular}{| l | l |}
\hline
\textbf{Model Changes} & \textbf{Dev WER (\%)} \\\hline
No changes (full global attention model) & 21.81 \\\hline
No sampling & 22.05 \\
No dropout & 22.17 \\
No input feeding & 22.06 \\\hline
No attention & 22.98 \\
No attention (rev. unidirectional encoder) & 22.65 \\\hline
\# of LSTM units - 256 & 24.00 \\
\# of LSTM units - 50 & 32.70 \\\hline 
2-layer LSTM & 22.36 \\
1-layer LSTM & 23.67 \\\hline
Rev. unidirectional encoder & 22.12 \\
Rev. unidirectional encoder + GRU & 23.78 \\\hline
%1-layer + Uni-directional Encoder + GRU & 26.7\\\hline
\end{tabular}
\caption{Ablation study on CMUDict development set.}
\label{tab:ablation}
\end{table}
In order to measure the contributions of various components of our attention models, we performed an ablation analysis for the global attention model evaluated on the CMUDict development data.  The results are shown in Table~\ref{tab:ablation}.  As can be seen from the table, the removal of input feeding results in a very minor drop in performance. The use of regularization in the form of dropout and scheduled sampling also provides a minor boost. 
The importance of attention is reflected in almost a 1\% absolute drop in performance when attention is removed. As we will later see, this boost is more prominent for longer words. 

In addition, we also tested the effect of bidirectionality by comparing against a ``reverse unidirectional'' LSTM, which drops the bidirectionality and takes the input in reverse order.  
Interestingly, the reverse unidirectional non-attention model is better than its bidirectional counterpart. This might be due to how we initialize our decoder LSTM for the bidirectional case, where we use a linear combination of hidden states of the forward and backward encoder LSTMs.  
In fact, bidirectional LSTMs may not be needed for a task with such short input sequences, as is reflected in the very minor drop in performance for the reverse unidirectional global attention model.

A large number of LSTM hidden units is crucial, as can be seen from the almost 11\% absolute drop in performance when we reduce it to 50.  The use of a 3-layer stacked LSTM is also justified, as a smaller number of stacked layers results in significant performance losses.  The importance of LSTM units can be gauged by the performance drop when they are replaced by Gated Recurrent Units (GRUs)~\cite{gru} in a reverse unidirectional global attention model.

\section{Analysis}
In this section, we analyze the models and their errors on the CMUDict data set.
\subsection{Error Comparison Based on Word Length}
\begin{figure}
\centering
\includegraphics[width=0.45\textwidth]{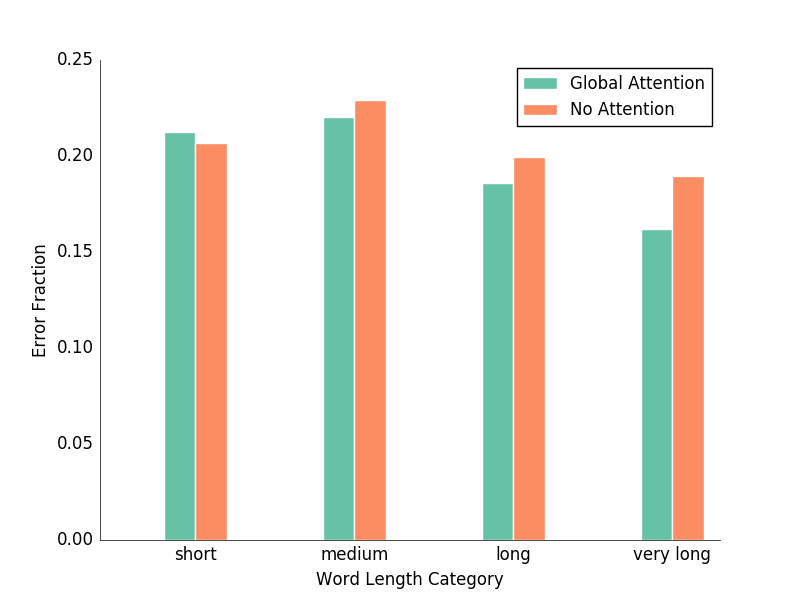}
\caption{Comparison of the word error rates of the global attention and no-attention model as a function of word length.}
\label{fig:err_len}
\end{figure}

We compare the errors made by the global attention model and no-attention model as a function of word length. For this purpose, we categorize the word length into 4 categories: \emph{short} (length $\le$ 6), \emph{medium} (length 
$\in \left\{7, 8\right\}$), \emph{long} (length 
$\in \left\{9, 10\right\}$), \emph{very long} (length 
$\ge 11$). 
Figure~\ref{fig:err_len} provides the results of this comparison. There are two main observations that can be made from the plot: (i) Both models make fewer errors on longer words; this is in some ways counter-intuitive, but is perhaps because the longer a word is, the more opportunity the model has to discern certain important global properties such as its origin. (ii) The global attention model has an even larger advantage over the no-attention model for longer words.  The second observation is quite intuitive:  The no-attention model is forced to represent all of the information about a word with a single hidden vector, which naturally fails for longer sequences.

\subsection{Phoneme Error Comparison Between Attention Models}
We next analyze the per-word PER, defined as the proportion of phoneme errors per word.  We are interested to analyze the distribution of phoneme errors across words.  That is, how often are word errors due to very minor phone errors, and how often are they due to a catastrophic failure?  
We consider this question by analyzing all of the word errors made by the global and local-$p$ attention models.  We categorize the per-word PER into 4 categories: \emph{small} ($\le$ 10\% of phones are incorrect), \emph{medium} (10-20\% incorrect), \emph{large} (20-30\%) and \emph{very large} ($\ge$ 30 \%).

We find that the numbers of \emph{small} and \emph{medium} errors are comparable for the two types of attention models.  However, the local-$p$ makes many more \emph{large} and \emph{very large} errors.  As might be expected, the local-$p$ attention model depends a great deal on the predicted alignment position;  if that prediction is incorrect, this can lead to cascading error effect and make it hard for the model to recover.

\subsection{Error Classification}
\begin{savenotes}
\begin{table*}[t]
  \centering
\begin{tabular}{|c | c c c c|}
\hline
 		&  {\bf Foreign Origin Names}  &  {\bf Under/Over Conversion} & {\bf Abbreviations} & {\bf Wrong Ground Truth} \\\hline
Word 			 		  &  QUIXOTE		& 	KITTIWAKE	 		& BLVD & 						ANALYSES\footnote{In the updated cmudict-0.7b, the model prediction is one of the pronunciations while the given ground truth has been removed}  \\
Ground Truth 	 &  K IY HH OW T IY & K IH T AH W EY K  & B UH L AH V AA R D  & AH N AE L AH S IY Z \\
Prediction 		 &  K W IH K S OW T &  K IH T AH W W K K   & B L AH D 				& AE N AH L AY Z AH Z \\\hline
Word 		 & MACIOCE  	&  	LASTS		& JNA 	& COMME\emph{RIC}AL \\
Ground Truth & M AA CH OW CH IY   & L AE S T S  & JH EY EH N EY & K AH M ER SH AH L \\
Prediction & M AH S IY OW S   & L AE S & N AH & K AH M EH R AH K AH L\\\hline
\end{tabular}
\caption{Examples of errors made by the global attention model.  Note that \textit{commerical} is a typo in the data set.}
\label{tab:error2}
\end{table*}
\end{savenotes}

The low PER of the global attention model means that even most of the incorrect predictions are still very close to the ground truth.  We analyze the cases where the prediction is still very far from the ground truth, in order to understand the kinds of phenomena that still need to be addressed.

We consider the following four most common categories of errors, also illustrated in Table \ref{tab:error2}:
\begin{itemize}
\item \textbf{Foreign Origin Names:} These examples can be very challenging, and at the same time fairly common as well. To give an idea of their importance, among the 13 words in the CMUDict development set with a phonetic edit distance of at least 4 between the ground truth and prediction, 11 of them were from this category.  In the examples presented in Table \ref{tab:error2}, the word \emph{QUIXOTE} is of Spanish origin and \emph{MACIOCE} is an Italian last name. 
\item \textbf{Under/Over Production:} In this type of error, the attention model either fails to attend to a character or attends to the same one too many times. For example, for the word {\it KITTIWAKE}, the model outputs \pron{K IH T AH W W K K}, as shown in Table \ref{tab:error2}, which seems to be due to translating the characters {\it W, K} twice. 
Conversely, for {\it LASTS} the model fails to convert the characters {\it T, S} at all, resulting in the output prediction \pron{L AE S}. This problem has also been observed in neural machine translation, and there has been some recent work to address it \cite{coverage}. The key idea is to maintain a \emph{coverage vector} during decoding which indicates the part of source that has been translated.  This approach may be a useful avenue for future work on G2P as well.

\item \textbf{Abbreviations:} This category is extremely difficult, but a bit rarer than the Foreign Names category.  As shown in the table, for the word \textit{BLVD}, which is an abbreviation of \textit{BOULEVARD}, the model outputs [ B L AH D ].  Some of the abbreviations in the data set are extremely challenging, requiring word knowledge as in this case, and the attention model seemed to struggle a lot with them.
\item \textbf{Wrong Ground Truth:} This is a very rare category but interestingly, for all of the instances of this type of error made by the attention model, the model's prediction matches the corrected version in the updated version of CMUDict.  For instance, for the word \textit{STACIE}, the model correctly outputs [ S T EY S IY ] but the ground truth is labeled as [ S T AE K IY ]. Another example is the misspelled word \textit{COMMERICAL}, shown in Table \ref{tab:error2}, with letters \textit{I, C} interchanged. Even for this non-existent word, the attention model outputs quite a convincing pronunciation. These errors suggest the need to update the version of CMUDict used in G2P evaluation for more meaningful results. 
\end{itemize}

\subsection{Phoneme Embedding Visualization}
\begin{figure}[t]
\centering
\includegraphics[width=0.45\textwidth]{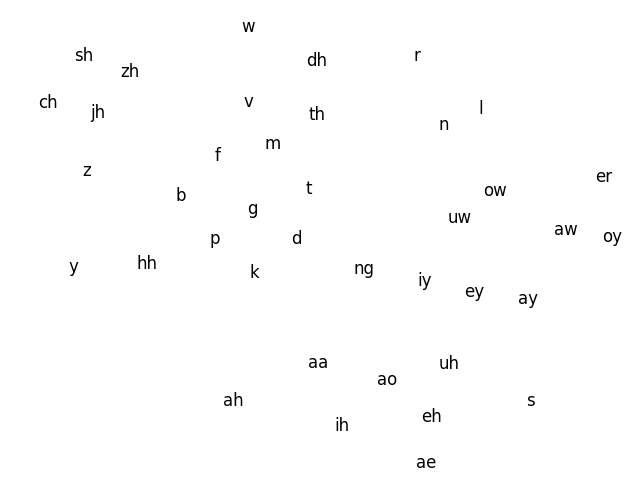}
\caption{Learned phoneme embeddings visualized using t-SNE.}
\label{fig:phone_emb}
\end{figure}

Most of the model parameters do not have a clear meaning.  However, the learned phone embeddings should be meaningful; for example, similar phones should be assigned similar embeddings and vice versa.  In order to confirm this, we plot in Figure~\ref{fig:phone_emb} a two-dimensional visualization, generated with t-SNE \cite{tsne}, of the phone embeddings learned by the global attention model.

The embeddings for the most part behave as expected.  The phones are broadly clustered according to manner classes: vowels, diphthongs, fricatives, and stops generally each have their own portion of the space (with some intermingling of stops and fricatives), and diphthongs form a tight sub-group within the vowel region.  Voiced/voiceless pairs of stops and fricatives are grouped close together, and vowels are placed in reasonable locations corresponding their height/frontness.  Nasals and semivowels are the least well-clustered, as might be expected.

\section{Conclusion}
In this work, we have applied an attention-enabled \seqtoseq{} model for the problem of grapheme-to-phoneme conversion. Our attention models achieve state-of-the-art results on three standard data sets, thus eliminating the need for explicit alignments which have been used by the best previous approaches. We compare several attention models and find that global attention is a reasonable choice given the short source sequence length for the G2P task. We also show that a simple monotonic alignment-based local attention is also well-suited for this task. Exploration of additional local attention models may therefore be a good direction for future work.

Our error analysis indicates that foreign names are one of the biggest remaining sources of error. 
This issue has been studied in some prior work~\cite{waxmonsky-reddy-naacl2012}, but has not been explicitly addressed in current state-of-the-art methods.  It may therefore be helpful to consider extending approaches such as ours with joint models of word origin and pronunciation.

\bibliographystyle{IEEEbib}
 \bibliography{g2p}

\begin{thebibliography}{10}

\bibitem{Chen}
Stanley~F. Chen,
\newblock ``Conditional and joint models for grapheme-to-phoneme conversion,''
\newblock in {\em Eurospeech}, 2003.

\bibitem{Bisani}
Maximilian Bisani and Hermann Ney,
\newblock ``Joint-sequence models for {G}rapheme-to-{P}honeme {C}onversion,''
\newblock {\em Speech Commun.}, vol. 50, no. 5, pp. 434--451, May 2008.

\bibitem{m2malignment}
Sittichai Jiampojamarn, Grzegorz Kondrak, and Tarek Sherif,
\newblock ``Applying {M}any-to-{M}any {A}lignments and {H}idden {M}arkov
  {M}odels to {L}etter-to-{P}honeme {C}onversion,''
\newblock in {\em NAACL-HLT}, 2007.

\bibitem{hochreiter1997lstm}
Sepp Hochreiter and J\"{u}rgen Schmidhuber,
\newblock ``Long {S}hort-{T}erm {M}emory,''
\newblock {\em Neural Comput.}, vol. 9, no. 8, pp. 1735--1780, Nov. 1997.

\bibitem{Rao}
Kanishka Rao, Fuchun Peng, Hasim Sak, and Fran{\c{c}}oise Beaufays,
\newblock ``Grapheme-to-phoneme conversion using {L}ong {S}hort-{T}erm {M}emory
  {R}ecurrent {N}eural {N}etworks,''
\newblock in {\em ICASSP}, 2015.

\bibitem{Yao}
Kaisheng Yao and Geoffrey Zweig,
\newblock ``Sequence-to-{S}equence {N}eural {N}et {M}odels for
  {G}rapheme-to-{P}honeme {C}onversion,''
\newblock in {\em Interspeech}, 2015.

\bibitem{bahdanau}
Dzmitry Bahdanau, Kyunghyun Cho, and Yoshua Bengio,
\newblock ``Neural {M}achine {T}ranslation by {J}ointly {L}earning to {A}lign
  and {T}ranslate,''
\newblock {\em CoRR}, vol. abs/1409.0473, 2014.

\bibitem{LAS}
William Chan, Navdeep Jaitly, Quoc~V. Le, and Oriol Vinyals,
\newblock ``Listen, {A}ttend and {S}pell,''
\newblock {\em CoRR}, vol. abs/1508.01211, 2015.

\bibitem{SAT}
Kelvin Xu, Jimmy Ba, Ryan Kiros, Kyunghyun Cho, Aaron Courville, Ruslan
  Salakhudinov, Rich Zemel, and Yoshua Bengio,
\newblock ``Show, {A}ttend and {T}ell: {N}eural {I}mage {C}aption {G}eneration
  with {V}isual {A}ttention,''
\newblock in {\em ICML}, 2015.

\bibitem{Galescu}
Lucian Galescu and James~F. Allen,
\newblock ``Bi-directional conversion between graphemes and phonemes using a
  joint n-gram model,''
\newblock in {\em 4th {ITRW} on Speech Synthesis}, 2001.

\bibitem{Wang}
Dong Wang and Simon King,
\newblock ``Letter-to-{S}ound {P}ronunciation {P}rediction using {C}onditional
  {R}andom {F}ields,''
\newblock {\em {IEEE} Signal Process. Lett.}, vol. 18, no. 2, pp. 122--125,
  2011.

\bibitem{CTC}
Alex Graves, Santiago Fern\'{a}ndez, and Faustino Gomez,
\newblock ``Connectionist temporal classification: Labelling unsegmented
  sequence data with recurrent neural networks,''
\newblock in {\em ICML}, 2006.

\bibitem{seq2seq}
Ilya Sutskever, Oriol Vinyals, and Quoc~V. Le,
\newblock ``Sequence to {S}equence {L}earning with {N}eural {N}etworks,''
\newblock in {\em NIPS}, 2014.

\bibitem{graves2013generating}
Alex Graves,
\newblock ``Generating {S}equences with {R}ecurrent {N}eural {N}etworks,''
\newblock {\em CoRR}, vol. abs/1308.0850, 2013.

\bibitem{lstm_eqn}
Wojciech Zaremba, Ilya Sutskever, and Oriol Vinyals,
\newblock ``Recurrent {N}eural {N}etwork {R}egularization,''
\newblock {\em CoRR}, vol. abs/1409.2329, 2014.

\bibitem{Vinyals}
Oriol Vinyals, Lukasz Kaiser, Terry Koo, Slav Petrov, Ilya Sutskever, and
  Geoffrey~E. Hinton,
\newblock ``Grammar as a {F}oreign {L}anguage,''
\newblock in {\em NIPS}, 2015.

\bibitem{Luong}
Thang Luong, Hieu Pham, and Christopher~D. Manning,
\newblock ``Effective {A}pproaches to {A}ttention-based {N}eural {M}achine
  {T}ranslation,''
\newblock in {\em EMNLP}, 2015.

\bibitem{adam}
John Duchi, Elad Hazan, and Yoram Singer,
\newblock ``Adaptive {S}ubgradient {M}ethods for {O}nline {L}earning and
  {S}tochastic {O}ptimization,''
\newblock {\em JMLR}, vol. 12, pp. 2121--2159, July 2011.

\bibitem{srivastava2014dropout}
Nitish Srivastava, Geoffrey~E Hinton, Alex Krizhevsky, Ilya Sutskever, and
  Ruslan Salakhutdinov,
\newblock ``Dropout: a simple way to prevent neural networks from
  overfitting,''
\newblock {\em JMLR}, vol. 15, no. 1, pp. 1929--1958, 2014.

\bibitem{sampling}
Samy Bengio, Oriol Vinyals, Navdeep Jaitly, and Noam Shazeer,
\newblock ``Scheduled {S}ampling for {S}equence {P}rediction with {R}ecurrent
  {N}eural {N}etworks,''
\newblock {\em CoRR}, vol. abs/1506.03099, 2015.

\bibitem{tensorflow}
Mart\'{\i}n Abadi, Ashish Agarwal, Paul Barham, Eugene Brevdo, Zhifeng Chen,
  Craig Citro, Greg~S. Corrado, Andy Davis, Jeffrey Dean, Matthieu Devin,
  Sanjay Ghemawat, Ian Goodfellow, Andrew Harp, Geoffrey Irving, Michael Isard,
  Yangqing Jia, Rafal Jozefowicz, Lukasz Kaiser, Manjunath Kudlur, Josh
  Levenberg, Dan Man\'{e}, Rajat Monga, Sherry Moore, Derek Murray, Chris Olah,
  Mike Schuster, Jonathon Shlens, Benoit Steiner, Ilya Sutskever, Kunal Talwar,
  Paul Tucker, Vincent Vanhoucke, Vijay Vasudevan, Fernanda Vi\'{e}gas, Oriol
  Vinyals, Pete Warden, Martin Wattenberg, Martin Wicke, Yuan Yu, and Xiaoqiang
  Zheng,
\newblock ``{TensorFlow}: Large-scale machine learning on heterogeneous
  systems,'' 2015.

\bibitem{gru}
Kyunghyun Cho, Bart van Merrienboer, {\c{C}}aglar G{\"{u}}l{\c{c}}ehre, Dzmitry
  Bahdanau, Fethi Bougares, Holger Schwenk, and Yoshua Bengio,
\newblock ``Learning {P}hrase {R}epresentations using {RNN} {E}ncoder-{D}ecoder
  for {S}tatistical {M}achine {T}ranslation,''
\newblock in {\em EMNLP}, 2014.

\bibitem{coverage}
Zhaopeng Tu, Zhengdong Lu, Yang Liu, Xiaohua Liu, and Hang Li,
\newblock ``Coverage-based neural machine translation,''
\newblock {\em CoRR}, vol. abs/1601.04811, 2016.

\bibitem{tsne}
Laurens van~der Maaten and Geoffrey Hinton,
\newblock ``Visualizing data using t-{SNE},''
\newblock {\em JMLR}, vol. 9, no. Nov, pp. 2579--2605, 2008.

\bibitem{waxmonsky-reddy-naacl2012}
Sonjia Waxmonsky and Sravana Reddy,
\newblock ``G2{P} conversion of proper names using word origin information,''
\newblock Stroudsburg, PA, USA, NAACL-HLT, 2012, pp. 367--371.

\end{thebibliography}

\end{document}